# Characterization of Human Balance through a Reinforcement Learning-based Muscle Controller

Kübra Akbaş[1], Carlotta Mummolo[2], Xianlian Zhou[1]*

1 Department of Biomedical Engineering, New Jersey Institute of Technology, Newark, NJ 07102
2 Department of Mechanics, Mathematics, and Management, Politecnico di Bari, Bari, Italy, 70125
*Corresponding author: alexzhou@njit.edu

**Abstract:** Balance assessment during physical rehabilitation often relies on rubric-oriented battery tests to score a patient's physical capabilities, leading to subjectivity. While some objective balance assessments exist, they are often limited to tracking the center of pressure (COP), which does not fully capture the whole-body postural stability. This study explores the use of the center of mass (COM) state space and presents a promising avenue for monitoring the balance capabilities in humans. We employ a musculoskeletal model integrated with a balance controller, trained through reinforcement learning (RL), to investigate balancing capabilities. The RL framework consists of two interconnected neural networks governing balance recovery and muscle coordination respectively, trained using Proximal Policy Optimization (PPO) with reference state initialization, early termination, and multiple training strategies. By exploring recovery from random initial COM states (position and velocity) space for a trained controller, we obtain the final BR enclosing successful balance recovery trajectories. Comparing the BRs with analytical postural stability limits from a linear inverted pendulum model, we observe a similar trend in successful COM states but more limited ranges in the recoverable areas. We further investigate the effect of muscle weakness and neural excitation delay on the BRs, revealing reduced balancing capability in different regions. Overall, our approach of learning muscular balance controllers presents a promising new method for establishing balance recovery limits and objectively assessing balance capability in bipedal systems, particularly in humans.

## 1. Introduction

Falls and subsequent injuries pose a significant health risk for the elderly and mobility-impaired populations. Poor balancing capabilities are the leading cause of falls in the elderly population, which reduces the overall quality of life of aging patients [1-3]. The injuries sustained by these patients can range from lower-body fractures, particularly in the hip, to head injuries, with falls being the leading cause of traumatic brain injuries [4]. Therefore, effective balance assessment and rehabilitation are critical components not only to health monitoring and injury prevention in mobility-impaired individuals, but also to the diagnoses of other serious underlying medical conditions. Since balance is maintained through a complicated network of physiological systems in the body, it is difficult to pinpoint a single origin causing deficiencies in patients and to assess balance through simple isolated measures. In most clinical environments, balance assessment is performed as a battery of balance exercises designed to evaluate the patient's ability to perform selected tasks. These exercises are often individually scored according to a specified rubric based on how well a patient can perform the activity, and the final cumulative score is used to categorize the subject into a specific level of balance ability (e.g., "high risk of falling"). For instance, the Berg Balance Scale indicates that a patient with a higher score would have better balance and a low risk of falling in their daily life; on the other hand, a low score would suggest that a patient has a higher risk of falling and would lead to further rehabilitation [5]. In addition to the Berg Balance Scale, other similar tests include the Mini BESTest [6], Balance Error Scoring System [7], Activities-Specific Balance Confidence Scale [8], Y Excursion Balance Test [9], and the Star Excursion Balance Test [9-11]. However, these presented balance assessment techniques can all be influenced by the subjectivity that is inherent within their observation-based scores; additional layers of subjectivity manifest themselves within the choice of test for a specific patient and the interpretation of the resulting scores. Moreover, clinical scores are typically applied to a battery of tests administered at different times (or pre-/post-intervention), hence monitoring the trend of the balance performance. With a single observation or test, typically, the status of the patient cannot immediately be inferred; thus, such tests alone are mostly not used for diagnosis. Since the problem of accurately assessing balance and recovery is multi-faceted, introducing objective and customized measures into the treatment plan can aid in providing better care to patients.

To create more objective assessment criteria, instrumented platforms and devices are often used to obtain kinematic or dynamic measures of a subject's posture, such as the Center of Pressure (COP) sway and sway velocity [12-14]; other measures include the Center of Mass (COM) sway in dynamic balance exercises [15-17]. Both COP- and COM-based approaches are related to the postural stability of a subject, by giving a measure of the subject's limit of dynamic balance to varying degrees of complexity. However, these measures alone are not directly predictors of falling, since they do not capture the whole-body and muscular dynamics, necessary to evaluate the specific limits of dynamic balance of the system [18]. On the other hand, approaches involving the partitioning of a state space of a legged system [19] have been explored, among which the COM state space has shown to be successful in characterizing the limits of dynamic balance of legged systems [20-23]. Based on the state space and general viability kernel concepts [23-25], a constrained optimization problem formulated using nonlinear programming was developed for the biomechanical analysis of balance using the COM state (position and velocity) [20, 26, 27], ultimately partitioning the state space into two sets: balanced and unbalanced states. Through this, a balance region (BR) can be constructed to describe a collection of COM states that serve as the necessary condition for a state of the given system to be balanced [20]. This region-based analysis provides a more comprehensive understanding of balance, when compared with the traditional COP- and COM-based metrics, by providing an estimate of the subject-specific limits of balance that take into account relevant physical constraints through complex dynamics and kinematics. Additionally, these regions can be employed to include metrics, like boundary and state margins, to provide a quantification of a subject's balancing ability [28, 29].

Although various balance assessment and tracking methods have been established or implemented, they are all limited by their reliance on only kinematics and dynamics information. While gross motion-based analyses can be useful in measuring the outcome, they lack the capability to examine the underlying muscular control mechanism that is physiological and has profound influence on balance in humans. The incorporation of muscle models can help account for physiological effects at the muscle level [30, 31], though recent efforts are being made to investigate the neural component as well [32]. Musculoskeletal (MSK) models help bridge the gap between multi-body dynamics approaches and physiology by linking joint actuation with individual muscle performance, leading to a deeper understanding of a person's motion. As the study of neuromusculoskeletal disorders expands, understanding the link between movement and muscular activity becomes increasingly important; disorders can affect individual muscle parameters, leading to a change in balance response, which may not be captured by a kinematic and dynamic approach. For instance, in Parkinson's Disease, there is a notable difference in maximum muscle force generation, force variability, and muscle activation timings [33, 34]. Therefore, for a more in-depth and physiological relevant understanding of balance, the use of a MSK model to study balance is of particular importance.

Developing a real-time muscle-based balance controller for humans is an effective approach to understanding how humans maintain balance and interact with their environment (e.g., in response to posture or force perturbation). Unlike traditional robot balance controllers that command joint motors directly (typically one motor for each joint), balance control with muscles is much more complicated due to the redundancy in muscles and intricate physiological response of muscle neural commands. Traditional musculoskeletal control problems rely heavily on experimental data such as muscle electromyography (EMG) or captured motions [35, 36], mainly due to the redundancy in muscle control. More recently, researchers have been using deep reinforcement learning and neural networks to control neuromusculoskeletal models to learn complex human movement skills [37-42].

To encapsulate and address the complexities of balance, reinforcement learning (RL) has been explored as a way of implementing higher-order models while bypassing the need to define every aspect of a problem (e.g., specifying all constraints in optimization, path-planning in traditional control). As machine learning algorithms in general become ubiquitous across all fields, their use in understanding gait and locomotion continues to grow and adapt to various problems that were previously difficult to address, especially RL algorithms [43, 44]. In legged robotics, RL algorithms can be formulated based on the desired outcomes or goals, but learning in the joint action space tends to be the most common approach [42, 45-47]. In all cases, joint-level dynamics must be considered when designing an algorithm; on the other hand, some approaches have considered the inclusion of high-level goals, such as whole-body postural changes or the tracking of ground reference points [48-51]. Unlike traditional model predictive controllers [21, 52] or those that rely on ground reference points for control [25, 51, 53, 54], RL based controllers can account for many factors that comprise balance and its subsequent assessment. A MSK model employed within the RL framework enables it to incorporate musculoskeletal characteristics (e.g., muscle strength, neural excitation-activation delay) of individuals or patients. Through domain randomization of the MSK model [40, 41], RL trained controllers can be robust against perturbation and inherent modeling errors.

In this study, we employ a musculoskeletal model controlled by a balance controller trained through RL to study human balance. Such a robust muscle controller enables a fast forward dynamics approach that brings the human from a perturbed state to a balanced (e.g., upright) posture and to establish the BR through wide exploration of the COM state space. By studying the BR attainable by RL-trained muscle controllers, we can gain deeper insights on neuromuscular control of balance and the confounding factors, such as muscle weakness and neural delay, that contribute to the deterioration of balance in humans. This work aims to provide a novel and more intuitive approach towards constructing BRs for the characterization and assessment of balance in bipedal systems, particularly in humans.

## 2. Materials and Methods

A muscular controller for postural balance recovery is developed using RL and is implemented with a MSK model to assess stability in the COM state space. The RL training environment adapts the structure of two interconnected neural networks (trajectory mimicking and muscle coordination) proposed in the Muscle-Actuated Skeletal System (MASS) framework developed by Lee at al. [42], where MASS is designed to track experimental data collected from motion capture. The approach in this study utilizes a desired equilibrium state (static, upright posture in double stance) as the target during training, thus eliminating the need for tracking experimental data. Simultaneously, neuromusculoskeletal physics and balance-inspired rewards are formulated in this work for the RL algorithm to effectively guide the learning process. After training, the controller is tested iteratively at varying initial states to generate a balance stability region in the COM state space. The presented RL-based controller allows for the control of individual muscles and can instantaneously respond to any state of the human to bring the state to equilibrium.

### 2.1. Musculoskeletal Model and Whole-Body Dynamics

A 2D MSK model (Figure 1) with 9 bilateral muscles (totaling 18 muscles) and 10 degrees of freedom (3-DOF planar pelvis joint, 1-DOF lumbar joint, and left/right symmetric 1-DOF hip, knee, and ankle joints) is adapted from the gait10dof18musc model available in the OpenSim repository [55]. In this work, the lumbar joint is locked to focus the analysis on the lower-limb joints and contributing muscles around these joints; however, the motion of lumbar joint can be allowed in this general framework for other types of analysis. For each foot, three contact spheres are positioned to establish contact with the ground. Specifically, one sphere is located on the heel, while the other two spheres are positioned near the toe joint. The heel contact sphere is positioned 4.9 cm behind the ankle joint along the anteroposterior (AP) direction, while the toe spheres are situated 15 cm in front of the ankle joint in the AP direction; the foot COM is located 5.1 cm in front of the ankle joint. When the model is in an upright standing position, the height of the COM ($H_{COM}^0$) is measured to be 98.8 cm. In the original gait10dof18musc model, the Millard muscle model with elastic tendon was used [56]. Nonetheless, a muscle model with rigid tendon is computationally much faster than its elastic counterpart and can achieve similar accuracy, especially for muscles with a relatively small ratio (< 1) of tendon slack length to muscle optimum length [56-58]. For the sake of efficiency, we implement a muscle model that is similar to the one implemented in MuJoCo [59], but with the addition of the pennation angle effect, which in theory is equivalent to the rigid-tendon muscle model presented in [56]. The physical parameters (such as fiber length, maximum muscle force) of each muscle are loaded from the OpenSim model without modifications.

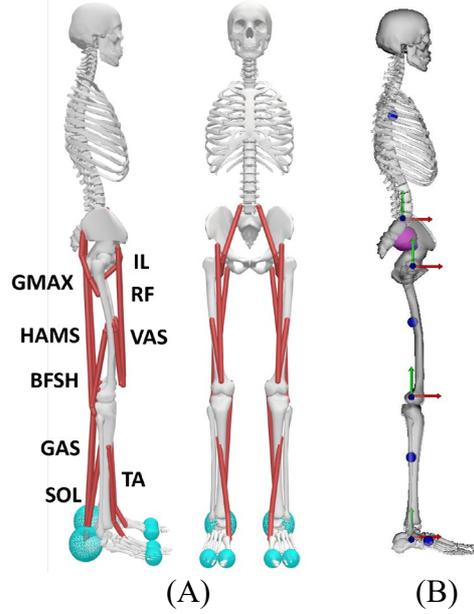

(A)          (B)

**Figure 1: A)** Musculoskeletal model with 18 muscles: gluteus maximus (GMAX), iliopsoas (IL), hamstrings (HAMS), rectus femoris (RF), vasti (VAS), biceps femoris short head (BFSH), gastrocnemius (GAS), soleus (SOL), and tibialis anterior (TA). **B)** Joint axes for ankle, knee, hip, and lumbar are presented in red (x), green (y), and blue (z) axes with the z-axis as the rotational axis for all joints. COM locations of individual bodies (foot, shank, thigh, pelvis, torso) are presented as blue spheres on the musculoskeletal model. The whole-body COM, which largely obscures the pelvis COM due to their near coincidence, is also presented in pink.

The muscle model takes the muscle force-length-velocity relations and fiber pennation angle into consideration when determining muscle force [56, 57], which is shown as follows:

$$F = F_{max} \cdot [a \cdot F_L(l) \cdot F_V(\dot{l}) + F_p(l)] \cdot cos(\alpha) \qquad (1)$$

where $F_{max}$ is the muscle-specific maximum isometric muscle fiber force, $a$ is the muscle activation ranging from 0 to 1, $\alpha$ is the fiber pennation angle, and $l$ is the normalized muscle length with respect to the optimal fiber length. Additionally, $F_L(l)$ and $F_V(\dot{l})$ are the normalized force-length and force-velocity curves, respectively. $F_p(l)$ represents the normalized passive force-length relationship.

The muscle activation ($a$) is governed by a first order excitation-activation dynamics equation:

$$\dot{a} = \frac{u-a}{\tau(u,a)} \qquad (2)$$

where $u$ is the muscle excitation (motor command or the control signal from the muscle network output) and $\tau$ is the delay time, which is computed as [56]:

$$\tau(u,a) = \begin{cases} \tau_{act}(0.5 + 1.5a) & u - a > 0 \\ \tau_{deact}/(0.5 + 1.5a) & u - a \leq 0 \end{cases} \qquad (3)$$

where $\tau_{act}$ and $\tau_{deact}$ are muscle activation and de-activation time constants with defaults of $(0.01, 0.04)$. The delay between muscle excitation and muscle activation can be interpreted as the time required for the excitation signal to propagate from the motor neurons to the muscle fibers and for the subsequent physiological processes to occur, resulting in muscle contraction. This dynamics equation is solved through integration, while both excitation and activation are clamped within [0,1].

The human musculoskeletal dynamics is mapped in the joint space, where it is governed by the Euler-Lagrangian equations using generalized coordinates:

$$M(q)\ddot{q} + c(q,\dot{q}) = J_M^T F_M + J_{ext}^T F_{ext} \qquad (4)$$

where $q$, $\dot{q}$, $\ddot{q}$ are the vectors of joint angles, angular velocity, and angular accelerations, respectively. $F_{ext}$ is the $R^3$ vector of external forces (such as contact forces) and $F_M$ is the $R^n$ vector of muscles forces ($n$ is the number of muscles) that depends on the muscle activation vector $a = (a_1, a_1, \cdots a_n)$. $M(q)$ is the generalized mass matrix, and $c(q,\dot{q})$ is the generalized bias force accounting for the Coriolis and gravitational forces. $J_M$ and $J_{ext}$ are the Jacobian matrices which map the muscle and external forces into generalized joint torques.

Since the muscle force is linear with respect to the activation as indicated in Eq. (1), the muscle forces are computed as:

$$F_M = \frac{\partial F_M}{\partial a} a + F_M(0) \qquad (5)$$

The corresponding generalized joint torques from muscles are defined by:

$$J_M^T F_M = J_M^T \left( \frac{\partial F_M}{\partial a} a + F_M(0) \right) \qquad (6)$$

When implemented in the RL framework, the dynamics of the musculoskeletal model are integrated using a forward dynamics approach with muscle excitations provided by the muscle coordination neural network. During the forward simulations, kinematic constraints such as joint limits are enforced and contact forces within the friction cone are solved with a linear complementarity problem (LCP) formulation, using the open-source DART simulation environment [60].

**2.2. Reinforcement Learning Framework**
The goal of the RL framework is to train the MSK model to recover balance from random initial states in the COM space, which are indirectly imposed by the angular position and velocity of the ankle joint. This generates a controller that takes as input the human body state information, predicts desired joint angles and transforms them into desired joint torques through PD control, and outputs muscle excitations as the control command for the physical MSK simulation environment (Figure 2). After the training phase, the controller is tested with many random initial states for balance recovery, which are classified as either successful or unsuccessful depending on if falling or foot movement happens, and the successful states are used to calculate the system's BR (Figure 3).

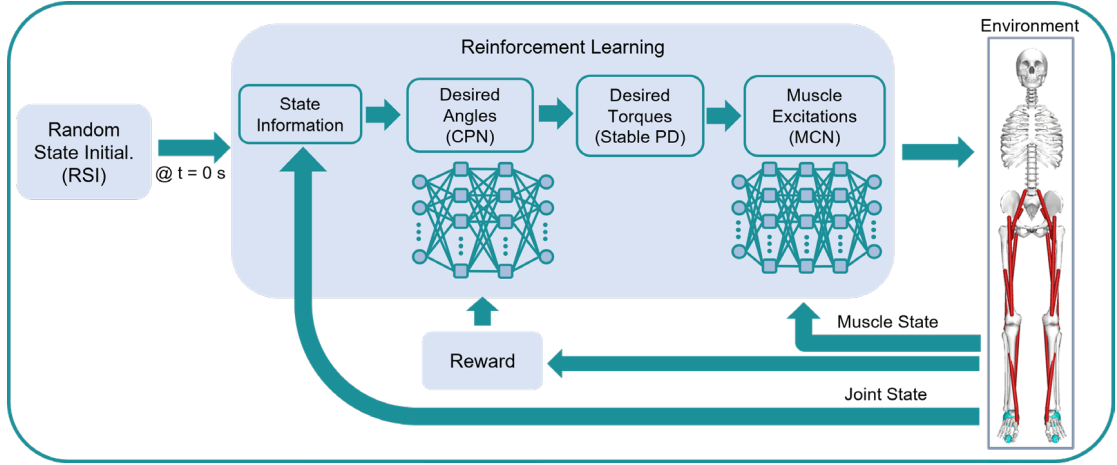

**Figure 2:** Overall control and RL framework. A random initial state is fed into the algorithm at the start of each episode and two neural networks (CPN and MCN) are used to control the MSK model. The RL rewards are computed to update the CPN, and a supervised loss function is used to update the MCN. The random initial state is given as a selected ankle position and velocity determined by the Reference State Initialization algorithm explained in Algorithm 1.

The learning environment is the dynamic simulator of the MSK model interacting with the ground. Control of the environment is achieved through a combination of two multilayer perceptron (MLP) neural networks (Figure 2): one is the control policy network (CPN) and the other is the muscle coordination network (MCN). The agent (CPN) takes as input the human body state information (the aggregations of the 3D positions and linear velocities of the COMs of all bones), then outputs the desired joint angles as the action. The desired joint angles are subsequently transformed to desired joint torques ($\tau_d$) through a stable Proportional-Derivative (PD) control [61]. The PD gains, $k_p$ and $k_v$, are set to 300 and $\sqrt{2 \times 300}$, respectively. Desired joint torques represent the target for muscle activation computation through the MCN. For apparent symmetric scenarios of the lower limbs, the dimension of the CPN's action can be reduced to a half of the number of controlled joints (bilateral hip, knee, and ankle joints) and transitioned back to the full dimension when fed into the MCN.

Separating joint motion control and muscle excitations into two networks allows the two networks to learn and operate at different frame rates. The joint network learns at a low frame rate (e.g., 30Hz) while the muscle network learns and operates at the rate of forward dynamics simulation (e.g., 600Hz). In addition, the joint network control policy $\pi_\theta(a|s)$ is a stochastic policy, whereas the muscle network is a deterministic policy that is learned through regression via supervised learning. The two networks mutually depend on each other and collaboratively interact with each other to achieve maximum rewards in RL.

*Muscle Coordination Network (MCN):* The neural network used for learning muscle excitations is a deterministic policy $a = \pi_\psi(\tau_d, s_{muscle})$, where the network parameters $\psi$ are learned through regression by supervised learning. The muscle network is set up as a MLP network with 3 hidden layers (n = 512, 256, 256 nodes) and the loss function for training is:

$$Loss(a(\psi)) = \mathbb{E}\left[\left\|\boldsymbol{\tau}_d - J_M^T\left(\frac{\partial \boldsymbol{F}_M}{\partial \boldsymbol{a}}\boldsymbol{a}(\psi) + \boldsymbol{F}_M(\boldsymbol{0})\right)\right\|^2 + w_{reg}\|\boldsymbol{a}(\psi)\|^2\right] \quad (7)$$

The first term is used to minimize the discrepancy between the desired torques and the muscle-produced joint torques under the predicted activation $\boldsymbol{a}(\psi)$. The second term is a regularization for large muscle activations. To enforce normalized muscle activations within [0,1], a bounded activation function is used for the output (i.e., Tanh function followed by a ReLU function). The MCN predicted $\boldsymbol{a}(\psi)$ is fed to the simulation environment as the muscle excitation instead of activation since the activation must obey Eq. (2).

***Control Policy Network (CPN):*** The CPN acts as the main RL agent controlling the MSK model's actions, based on its accumulated rewards. As the RL agent interacts with its environment, its actions are scored using a reward and the agent is updated based on the action's reward. At each time step $t$, the agent's state $s_t$ in the environment is observed and an action $a_t$ is selected according to its control policy $\pi_\theta(a_t|s_t)$ with $\theta$ being the weights and bias of the neural network. The control policy is learned by maximizing the discounted sum of reward ($r_t$):

$$\pi^\star = \underset{\pi}{\mathrm{argmax}}\, \mathbb{E}_{\tau \sim p(\tau|\pi)}\left[\sum_{t=0}^{T-1}\gamma^t r_t\right] \quad (8)$$

where $\gamma \in (0,1)$ is the discount factor, $\tau$ is the trajectory over time $\{(s_0,a_0,r_0),(s_1,a_1,r_1),\dots\}$, $p(\tau|\pi)$ is the likelihood of that trajectory $\tau$ under the control policy $\pi$, and $T$ is the horizon of an episode.

The RL framework utilized here is trained with the Proximal Policy Optimization (PPO) algorithm [62], which is a model-free policy gradient algorithm that is widely used for continuous control problems. The PPO agent samples interactions with the environment and optimizes a "surrogate" objective function. PPO updates the control policy's parameters ($\theta$) using the gradient of the expected return with respect to the parameters.

To ensure that the new, updated policy is close to the old policy, the PPO algorithm uses a trust region constraint with a probability ratio:

$$P_t(\theta) = \frac{\pi_\theta(a_t|s_t)}{\pi_{\theta,old}(a_t|s_t)} \quad (9)$$

where $\pi_\theta$ is the new control policy, $\pi_{\theta,old}$ is the previous control policy before the update. PPO uses an objective function that integrates the probability ratio through clipped probability ratios, which can provide a conservative estimate of the policy's performance. The "surrogate" objective function is defined as:

$$L(\theta) = \mathbb{E}_t\left[min(P_t(\theta)\hat{A}_t, clip(P_t(\theta), 1-\epsilon, 1+\epsilon)\hat{A}_t)\right] \quad (10)$$

where $\epsilon$ is a small positive number used to help constrain the probability ratio, and $\hat{A}_t$ is the advantage value that provides a measure of how good or bad a specific action is in a given state. The clipping function is used to prevent the policy from changing drastically and taking the minimum results by using the lower, pessimistic bound of the unclipped objective. The control policy can then be updated through maximizing the clipped discounted total reward in Eq. (10) using gradient ascent.

Through this PPO-based control policy, the agent learns to increase its reward by modifying the parameters $\theta$ of the network. This is implemented as a MLP network with 2 hidden layers of 256 nodes each (Figure 2). In the original MASS framework [42], the trajectory tracking component is used to learn specific joint motions needed to follow a motion from motion capture data. Here, rather than mimicking motions from experimental data, the framework is tailored to balancing tasks, where balance recovery motions are not provided or needed. Instead, a desired equilibrium state is provided as the target posture, for which the whole-body COM is situated right on top of the foot COM vertically. In addition, the present network does not require a phase variable, a number defined as the ratio between the current simulation time and the end time of reference motion, as an input.

*Reward Functions:* The reward function $r_t$ for the RL algorithm is designed to drive the MSK model to reach the target state by including a target posture reward $r_t^p$, a torque reward $r_t^{torque}$, a body upright reward $r_t^{upright}$, and an extrapolated center of mass (XcoM) [63] reward $r_t^{xcom}$, and is defined as follows:

$$r_t = w^p r_t^p + w^{torque} r_t^{torque} + w^{upright} r_t^{upright} + w^{xcom} r^{xcom} \quad (11)$$

where $w^p = 1.0$, $w^{torque} = 0.1$, $w^{upright} = 0.1$, and $w^{xcom} = 0.1$ are their respective weights. The target posture reward is designed to match the target posture with the actual joint angles:

$$r_t^p = \exp\left(-\sigma_p \sum_j \left\| \hat{q}_t^j - q_t^j \right\|^2\right) \quad (12)$$

where $j$ is the joint DOF (angle) index, $\hat{q}_t^j$ is the joint DOF value for the target standing posture, and $\sigma_p = 2.0$. Here we set the target hip and knee joint angles to be zero and the target ankle angle to be 5.56° degree such that the whole-body COM is situated on top of the foot COM.

The torque reward is included to help reduce energy consumption of the joints:

$$r_t^{torque} = \exp\left(-\sigma_{torque} \sum_j \left\| \tau_j \right\|^2\right) \quad (13)$$

where $\sigma_{torque} = 0.001$.

The upright posture reward is defined as

$$r_t^{upright} = \exp\left(-\sigma_{upright} \left\| p_{head}^x - p_{pelvis}^x \right\|^2\right) \quad (14)$$

where $p^x_{pelvis}$ is the $x$ (horizontal) position of the pelvis, $p^x_{head}$ is the $x$ position of a point on the head which equals to $p^x_{pelvis}$ when pelvis tilt angle is zero (i.e., torso is upright).

The XcoM reward is defined as:

$$r_t^{xcom} = \exp\left(-\sigma_{xcom}\|XcoM_t - XcoM_{target}\|^2\right) \tag{15}$$

where $XcoM_{target}$ is set as $x$ position of the foot's COM. The $XcoM$ concept is based on the region-based stability analysis of a Linear Inverted Pendulum (LIP) model [63], which is defined as:

$$XcoM = x + \frac{v}{\omega} \tag{16}$$

where $x$ is the horizontal COM position in the sagittal plane, $v = \dot{x}$ is the horizontal COM velocity, and $\omega = \sqrt{g/l}$ is the natural frequency of the LIP, where $g$ is the gravity constant and $l$ is the height of the LIP. For the human MSK model, $l$ is the whole-body COM height. In the literature, researchers often use Margin of Stability (MoS) to assess stability [63], which is defined as:

$$MoS = |BoS - XcoM| \tag{17}$$

where $BoS$ is the limits of the base of support (heel or toe) in the horizontal direction of the sagittal plane. For efficiency, instead of involving both heel and toe positions, we use the distance to $XcoM_{target}$ instead of the MoS for the reward.

***Training Strategy***: In the work by Peng et al. [64], two specific components, reference state initialization (RSI) and early termination, are identified to be critical for achieving highly dynamic motions from mimicking a reference motion. In their work, RSI is used to initiate the state of the model at the start of each episode by sampling the reference motion at a random time. By leveraging RSI, the agent can benefit from a diverse and informative distribution of states, which can effectively guide its training process. In this study, we also employ RSI for state initialization even in the absence of a reference motion. Initial joint space states for each training episode are randomly selected from normal distributions for both angular position (mean: $\mu_p$, SD: $\sigma_p$) and velocity (mean: $\mu_v$, SD: $\sigma_v$) of the ankle joint to encourage exploration of the COM state space. The other joint DOFs (hip and knee) are set to zeros for the initial posture, so that the model has straight legs in the beginning. During the simulations, all the joints are allowed to move except for the lumbar joint. By integrating RSI during training, it encourages exploration of the COM state space by exposing the controller to different initial states, which may range from relatively easy to highly challenging or even infeasible conditions across different episodes.

Algorithm 1 describes the procedure to generate randomized initial states with feet kept level with ground without movement (zero velocity), while the body incline and rotational velocity are randomized through ankle joint angle and velocity sampling from two normal distributions with prescribed means and SDs. We assume the knee and hip maintain their neutral posture (zero joint angles) at initialization and solve the translation velocity of the pelvis with a gradient descent minimization method that ensures the foot linear velocity is zero. The minimization procedure often converges within a few iterations.

---

**Algorithm 1:** Initial State Randomization

---

Set ($\mu_p, \sigma_p$) for ankle angular position

Sample ankle angle from a normal distribution ($\mu_p, \sigma_p$)

Set slope $s$, and ($\mu_v = s \times \mu_p, \sigma_v$)

Sample ankle angular velocity from a normal distribution with mean and SD ($\mu_v, \sigma_v$)

Set the pelvis tilt angle (and angular velocity) as opposite of the ankle angle (and angular velocity) to ensure leveled feet with zero rotational velocity; shift the pelvis position such that the feet remain at the same location.

Set $tolerance = 1e - 8, max_{iter} = 100$

for iteration =1,2, …, $max_{iter}$ do

    Set $error = 0, \Delta \dot{q} = 0$, step=0.25

    for each foot do

        $error += ||v_{foot}^{COM}||^2$  where $v_{foot}^{COM}$ is the linear velocity of the foot COM

        $\Delta \dot{q} -= 2.0 \times step \times J^T v_{foot}^{COM}$ where $J$ is the Jacobian of the foot COM linear velocity

    end for

    if $error < tolerance$, exit the for iteration

    Increment only the pelvis translation velocities with corresponding components in $\Delta \dot{q}$

end for

---

An early termination strategy offers an alternative means of shaping the reward function in order to discourage undesirable behaviors. Additionally, it can function as a curating mechanism that favors data samples that may be more pertinent to a given task. In this work, the following events are used as early termination conditions:
- Fall: pelvis height is lower than a threshold which is set to 0.8 m, corresponding to an ankle angle of 40°. At the upright posture, the pelvis height is 0.965 m and COM height is 0.9877 m.
- Foot slide: either foot moves more than 1 cm along the anteroposterior direction.
- Foot lift: either foot lifts more than 1 cm along the vertical direction.

During training, each episode is a musculoskeletal simulation that ends at 10 seconds unless it is terminated earlier.

The hyperparameters used in this study are presented in Table 1. These hyperparameters, along with the network layer depth and width, are either taken from literature [62, 65] or selected based on our empirical trials to obtain sufficient network representation capability and learning efficiency.

**Table 1:** Hyperparameters used for training.

| Parameters | Value | Parameters | Value |
|---|---|---|---|
| Discount Factor | 0.99 | Epochs | 10 |
| Policy Adam learning rate | $10^{-4}$ | Clip threshold | 0.2 |
| Batch Size | 128 | Memory buffer | 2048 |

Solving complex human movement and control problems with deep RL is prone to instability and frequently results in unfavorable local optima. To encourage convergence to a robust and natural human balance controller, a muscular balance controller is trained with three different approaches: 1) RSI with random initial ankle angle ($\sigma_p \neq 0$) and zero starting velocity ($\mu_v = \sigma_v = 0$), 2) RSI with random initial ankle angle and velocity ($\sigma_p \neq 0, \sigma_v \neq 0$), and 3) a two-step curriculum learning (CL) process by first training with zero starting velocity ($\mu_v = \sigma_v = 0$) and then continuing with non-zero random velocity ($\sigma_p \neq 0, \sigma_v \neq 0$). CL [66] has been used in literature to learn complex human and robot movement skills [37, 39, 67, 68]. CL enhances the learning process by breaking down a challenging task into multiple intermediate steps that are easier to learn, thereby facilitating progress towards a favorable direction. In the CL process, we start by learning to balance from an inclined state with zero initial velocity (by setting the initial velocity mean and SD $\mu_v = \sigma_v = 0$). After completion of the training, the best performing neural networks (with the highest reward or minimal loss) are used as the starting point of the next step, which uses non-zero velocity mean and SD ($\mu_v = s \times \mu_p$ with $s = -\omega$, $\sigma_v = 0.1\ rad/s$). $s$ is a slope factor that tilts the velocity mean based on the current sampled position $\theta_{ankle}$ and we use $\omega$ for it to follow the LIP balance region limits [29]. For the first two approaches, 50,000 iterations are used for each training course. For CL, 50,000 iterations are used for each training step. All training is performed on a Linux machine with Intel Xeon CPUs (2.30GHz) and a 16G Nvidia Quadro RTX 5000 GPU, and typical training time for 50,000 iterations is around 40 hours.

**2.3. Testing and Balance Region Generation**
For each training method, the learned balance controller is tested on the MSK model to examine its ability to regain balance from various initial states, as illustrated in Figure 3. During testing, the learned controller strives to drive the MSK model from a given random initial state to an upright balanced state without triggering the same terminating conditions that were used during training. To generate the BR with the trained controller, 10,000 simulations are performed with random initial ankle positions and velocities (as in method 2); this number is selected to ensure densely sampled initial states within reasonable computational time. The CPU time needed for BR generation was typically less than a few hours since each simulation terminates at 10s or less of the simulation time and the simulation was faster than real time (~5 times faster). The outcome of each simulation is recorded as "successful" if it runs to the end of the specified episode time (10 seconds) without triggering a termination condition. Otherwise, it is recorded as "unsuccessful". For every successful simulation, the corresponding initial COM position and velocity is stored as a point of the system's BR. A point-based BR (PBR) is generated from the collection of all successful initial COM states (a point cloud). However, during the course of dynamic balance recovery, a COM state trajectory may go outside of the PBR. Note that, all points on a successful trajectory (COM states at discrete times) lead to balance recovery at the end of the time interval. Therefore, these points along the successful trajectories should be considered as a part of the final BR. The PBR and (final) BR envelopes are generated from a point cloud using the alpha shape toolbox (https://github.com/bellockk/alphashape) in Python, which finds the bounding polygon of a point cloud. The point cloud's convex hull could also be considered, but was found to significantly overestimate the BR in the current work. On the other hand, the alpha shape bounding polygon can be concave but may not strictly encapsulate all the points in the cloud depending on the $\alpha$ parameter (representing a generalized disk of radius $1/\alpha$ used to draw the boundary edge [69]).

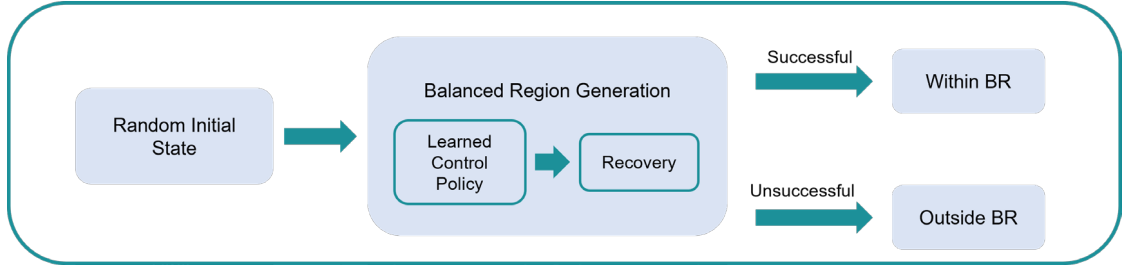

**Figure 3:** Schematic describing testing protocol for BR generation.

## 3. Results

The neural networks were trained using the three different methods outlined in the training strategy (RSI with random initial ankle angle and zero velocity, RSI with random initial ankle angle and velocity, and two-step CL). For the first method, the following parameters were used: $\mu_p = 0.09745\ rad$ corresponding to the target ankle angle (5.58°, plantarflexion), $\sigma_p = 0.1\ rad, s = 0, \mu_v = s \times \mu_p = 0,$ and $\sigma_v = 0$. For the second method, the same $\mu_p$ and $\sigma_p$ were used, while $s$ was set to $-\omega$ and $\sigma_v$ was set to 0.1 rad/s. As for the third (two-step CL) method, the training with the first method was repurposed as the first step and the second step involved using the best outcomes from the first step and performing another training using the identical parameters as the second method. The original rewards as well as the smoothed rewards obtained using a 5-point moving average for these training methods are presented in Figure 4. In the case of the first two methods, the reward exhibited a rapid increase from a very small value to over 200 within 5,000 iterations. It is noted that the overall reward from method 2 slightly surpassed that of method 1. As for the third method (CL), the second step reward started off at a high value. Further training led to incremental improvements in the maximum reward; however, these improvements occurred after a considerable number of iterations and the magnitude of each increase was not substantial.

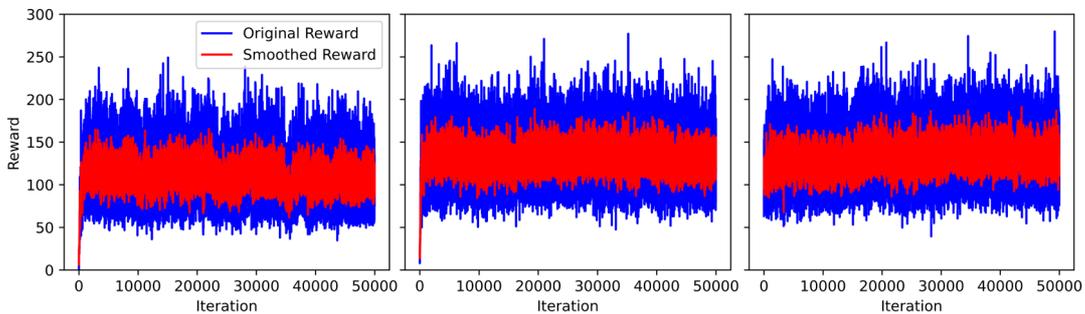

**Figure 4:** The rewards (original and smoothed rewards with 5-point moving average) for training with three different methods. Left: training with random initial ankle angle and zero starting velocity; middle: training with random initial ankle angle and velocity, and right: the second training step of the curriculum learning with random ankle angle and velocity.

Using the learned controllers from these three methods, we conducted tests to generate BRs with the approach outlined in Section 2.3. For each test, we ran 10,000 dynamic simulations with a random initial state sampling similar to the one used for the second training strategy (Algorithm 1). The successful and failed (early terminated) simulation trials of the three controllers are illustrated in Figure 5. The overall success rate of each controller is calculated as the number of successful simulation trials with respect to the total number of trials conducted, which resulted in 59.59%, 16.31%, 41.60%, for the three training methods, respectively. The envelopes of successful initial COM states (i.e., PBR) were generated from these successful points using alphashape. Within the PBRs, there are scattered failed points (blue markers) among the successful points (orange markers), most of which are not visible due to rendering. The PBR from method 1 contains 5959 successful and 74 failed points, the latter being mostly located near the bottom right corner where the PBR is concave. This indicates an "internal" success rate within the PBR equal to $5959/(5959 + 74) = 98.77\%$; smaller success rates within the PBRs for methods 2 and 3 are found (42.72% and 87.38%, respectively). The areas of the PBR for each method are also listed in the figure; methods 1 and 3 have comparable areas that are both over 45% larger than the area of method 2. The method 1 area is more convex, as compared to method 3, and has a higher success rate (98.77% vs 87.38%).

None of the PBRs includes the entire set of static equilibrium points (i.e., the states with zero COM velocity and COM position within the contact limit [29]), for which the COM ground projection coincides with the COP. The placement of the contact spheres in the current contact model, described in the previous section, limits the COP position within the anteroposterior range of [-4.9, 15] cm that is slightly smaller than typical foot size (since our frontal contact spheres are placed around the toe joints instead of the toe tips), excluding the possibility of the system to maintain a static posture with a COM projection beyond the toe joint. However, the set of static equilibrium points in the PBRs is even smaller than this range (19.9 cm), being reduced particularly in the anterior direction. Among the three methods, method 1 covers a much wider range of static equilibrium points.

As stated earlier, all the successful trajectories lead to balance recovery. Therefore, all the points along such trajectories should be included in the system's final BR. The trajectory-based BRs were constructed from a selection of successful COM trajectories in each of the three cases, as shown in Figure 5. For the sake of efficiency, only the trajectories originating from the boundary points of the PBR and from an additional 100 randomly chosen successful internal points were included in the generation of the final BR. This approach ensures computational efficiency while maintaining a representative sample of trajectories for analysis; nonetheless, a different approach could still be used to include more trajectories. The alpha shape envelopes for all cases were generated using α=15. Typically, a larger α value generates tighter and more concave envelopes, but in some cases, it may produce multiple disjointed envelopes. On the contrary, a smaller α value generates looser envelopes and likely overestimates the covered areas.

A trend similar to the PBRs is observed in the areas covered by the trajectory-based BR: method 2 covers the smallest area, whereas method 1 has a comparable area to that of method 3 but displays more convexity. On the other hand, contrary to the PBRs, the final BRs in all three methods mostly cover the entire set of equilibrium points, indicating that for any COM ground projection within the foot base of support there exists at least one balanced trajectory. This is possible because the states included in the BR belong to highly dynamic balancing trajectories, taking advantage of the presence of inertial forces [29].

For comparison, the expected analytical stability solution to the LIP model was calculated [63]. The bounds of the LIP region (i.e., limits of dynamic stability) were determined by $v_{max} = \omega(u_{max} - x)$ and $v_{min} = \omega(u_{min} - x)$ for the anterior and posterior directions, respectively, where $x$ is the COM $x$-position in the horizontal direction of the sagittal plane. The position ($x = 0$) occurs when the COM is aligned on top of the ankle joint. $u_{max} = 15\ cm$ and $u_{min} = -4.9\ cm$ are the locations of the toe and heel contact spheres and represent the maximum range of the COP displacement. The LIP stability limits, along with the zero COM velocity line, are plotted in conjunction with the BRs in Figure 5. It is observed that while all BRs largely fall within the analytical bounds, they can also exceed them in few instances; this is possible because the model in this study is multi-segmental and can exhibit some angular momentum strategy for balance recovery, unlike the LIP [29].

To investigate the outcome of the trained controllers on joint kinematics, we analyzed the joint angles of the 100 randomly selected successful trajectories for each of the three cases, as well as the mean final posture at the end of the 10-second simulations. The mean final postures of the MSK model achieved with the controller trained using method 1 resulted in a posture closest to the target posture (Figure 6). Additionally, the joint angles and COM positions of the final postures from the three training methods are analyzed against results from experimental data (Table 2). Experimental joint angles and COM positions were collected from standing trials performed in the BioDynamics Lab at NJIT under IRB approval #2212027868. Data was collected from 22 subjects (11 males and 11 females) who were instructed to stand straight with their eyes open and arms crossed against their chest. Joint and COM information were obtained through inverse and body kinematics, respectively, which were performed in OpenSim using the Hamner 2010 model [70, 71]. It is evident that the final posture resulting from method 1 is more aligned with a natural upright stance, when considering the final COM and joint states, whereas the final postures obtained from methods 2 and 3 exhibit increased forward inclination compared to the target posture. With the exception of hip flexion, method 1 closely resembles the standing posture obtained from experiments. Taking into account the success rate of PBRs (overall and internal), shape and coverage of BRs, and mean final posture, the controller trained using method 1 (random initial position and zero initial velocity of the ankle joint) appears to be more robust and performs the best.

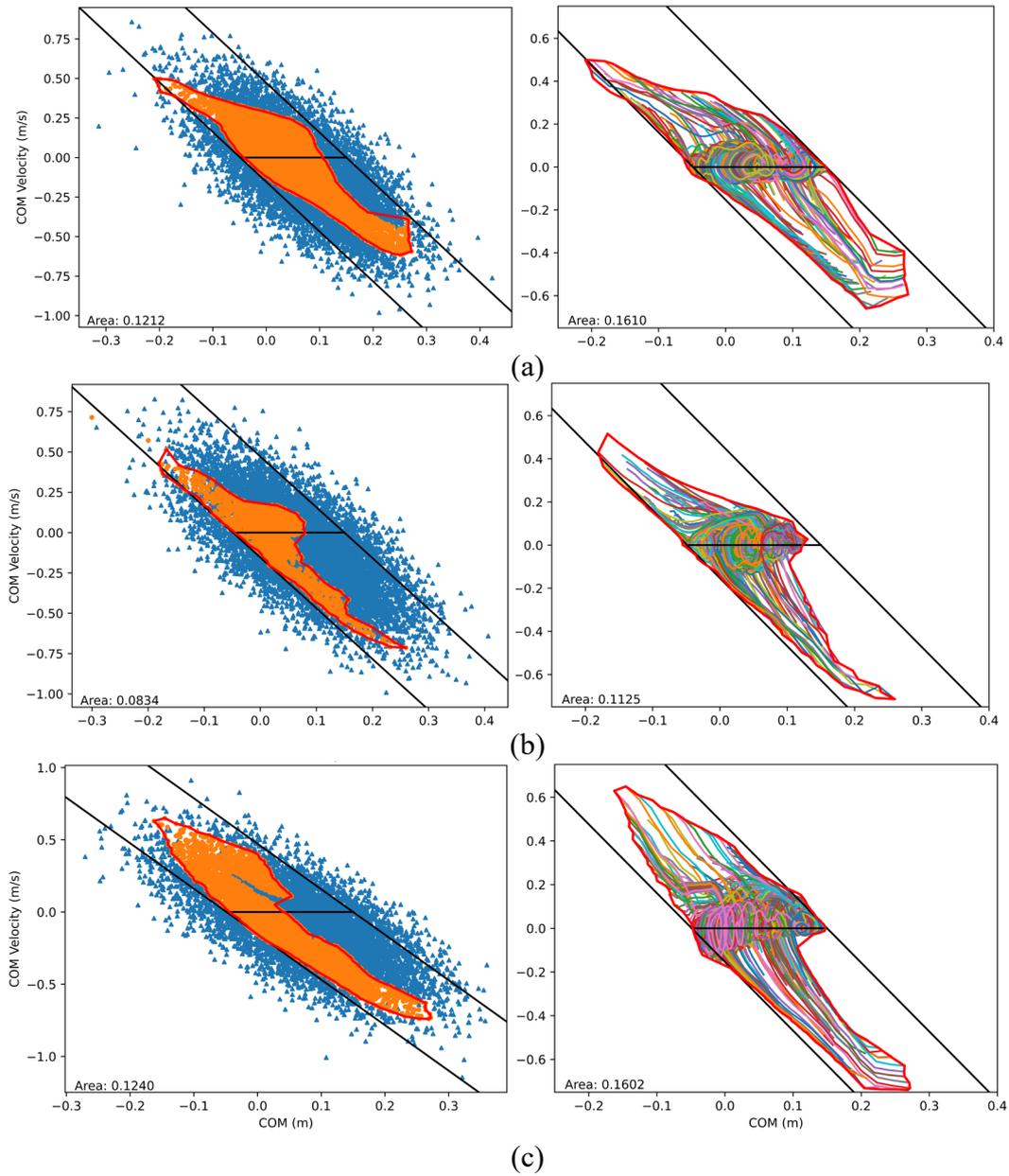

**Figure 5:** COM state space BRs (enclosed by the thick red curves) based on the testing of the learned controllers trained with (a) method 1 (zero starting velocity); (b) method 2 (non-zero velocity); and (c) method 3 (CL). Left: Initial COM states (points) of successful trials (orange markers) and unsuccessful trails (blue markers) and generated PBRs from the successful trials. Right: COM state trajectories of selected successful trials and generated final BRs.

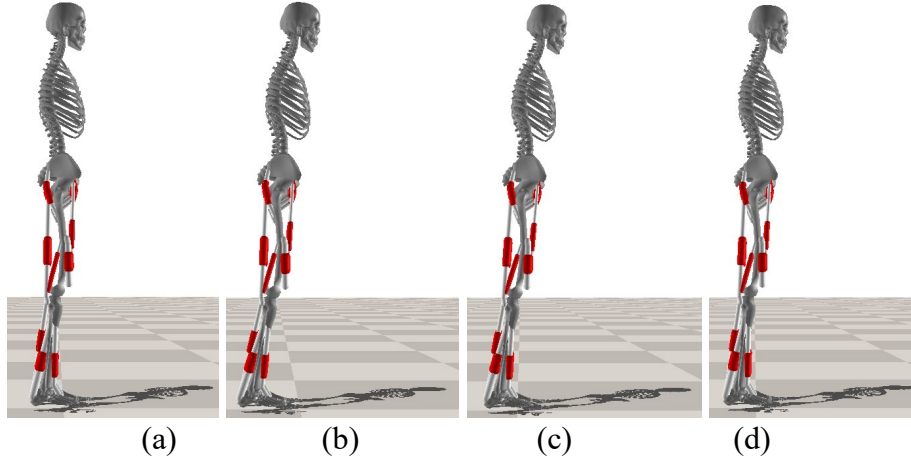

(a)          (b)          (c)          (d)

**Figure 6:** The mean postures at 10s for (a) method 1, COM position: 0.049 ± 0.023 m; (b) method 2, COM position: 0.108 ± 0.033 m; and (c) method 3, COM position: 0.104 ± 0.023 m; and (d) target posture, COM position: 0.051 m. In the figure, the muscle tendons are displayed as narrow white cylinders.

**Table 2:2** Mean and SD of the final postures from the three training methods. The angles are in degrees and the COM position is in meters.

|   | Method 1 | Method 2 | Method 3 | Experimental |
|---|---|---|---|---|
| Pelvis tilt | −3.388 ± 3.879 | −8.663 ± 2.782 | −9.162 ± 2.668 | −2.885 ± 3.059 |
| Hip angle | −0.606 ± 1.136 | 3.265 ± 1.493 | 1.896 ± 0.688 | −6.750 ± 3.543 |
| Knee angle | −1.436 ± 4.093 | −4.560 ± 1.070 | −0.261 ± 1.726 | 0.763 ± 4.327 |
| Ankle angle | 5.971 ± 1.541 | 10.421 ± 1.404 | 8.103 ± 0.500 | 5.289 ± 2.110 |
| COM Position | 0.049 ± 0.023 | 0.108 ± 0.033 | 0.104 ± 0.023 | 0.0302 ± 0.0155 |

*Negative pelvis tilt angle indicates leaning forward; negative hip angle indicates hip extension; negative knee angle indicates knee flexion; positive ankle angle indicates dorsiflexion.

**Balance Recovery from Forward and Backward Lean.** Using the trained controller from training method 1, balance recovery from forward and backward leans is analyzed. The RSI algorithm is used to obtain 100 random initial states; a mean ankle angle $\mu_p = 8°$ (close to the model's recovery limit in PBR) with a SD of $\sigma_p = 0.1°$ and zero velocity is set for forward lean tests and a mean ankle angle of $\mu_p = -1.45°$ (near its backward extreme) with a STD of $\sigma_p = 0.1°$ and zero velocity is set for the backward lean tests. In both tests, we ran muscle-controlled dynamic simulations to collect 100 successful trials. COM trajectories of these trials for balance recovery from forward lean, along with the time history profiles of the mean COM position and velocity, are presented in Figure 7. During balance recovery, the COM initially moves forward with a positive velocity, gradually decelerating until it reaches its furthest point where its velocity becomes zero. Subsequently, the COM retraces and passes the original starting position, moving toward the COM of the foot in the AP direction. The time history data reveals that the variations in motion increase over time, which can be attributed to two factors: 1) small variation in the starting position is cumulated during dynamic time integration; 2) the first control policy network is stochastic and employs a multivariate normal distribution for output, introducing additional variations during the simulation.

The time history profiles of the mean joint angles and muscle activations are displayed in Figure 8. The tilt angle of the pelvis determines the upper body inclination, given that the lumbar joint is locked. A negative tilt angle signifies forward inclination, whereas a positive angle indicates backward inclination. In the early stage of balance (within 1 second), the pelvis (and upper body) tilts further forward, while the ankle moves in the opposite direction (plantarflexion). Simultaneously, the knee extends further and reaches its limit, and the hip flexes, causing the leg to strive towards the more upright posture. After the initial 2 seconds of the simulation, the body gradually approaches the final converged state, exhibiting small oscillations. Regarding muscle activation, the iliopsoas muscle demonstrates the highest activation, peaking at a value close to 0.4, followed by the hamstrings, biceps femoris, and soleus. The tibialis anterior muscle exhibits greater activation in the first second of the simulation but has a relatively low level of activation throughout the remaining duration. On the contrary, the gastrocnemius, another ankle plantar flexor, displays minimal activation in the early stage of the balance recovery but intensifies its activation as the body attains a more upright posture toward the end of simulation. Unlike the tibialis, the gastrocnemius is a biarticular muscle, generating knee flexion torque when contracting, thus remaining mostly inactive during the initial balance phase when knee extends. Other muscles such as gluteus maximus, vasti, and rectus femoris are mostly inactive or have very low activation during forward lean balance recovery.

Figure 9 and Figure 10 present results for balance recovery from backward lean. Initially, the COM exhibits oscillations around its starting position for less than a second before accelerating with a positive velocity towards the COM of the foot in the AP direction. Around the three-second mark, the COM enters a state of oscillation around the final position, forming a circular pattern in the COM state space. The joint angle plot reveals an immediate dorsiflexion of the ankle and bending of the knee during the early recovery phase. However, the pelvis (and upper body) initially falls backward, leading to hip extension. Compared to the forward lean recovery, the joint angles display more pronounced oscillations in both magnitude and frequency. By analyzing the mean muscle activation, it is evident that the rectus femoris muscle exhibits the highest level of activation, reaching a peak slightly below 0.6. It is followed by the tibialis anterior muscle, hamstrings, and then the iliopsoas. For the entire duration, both the iliopsoas and hamstrings demonstrate low levels of activation (less than 0.1). The biceps femoris briefly exhibits low activation at the beginning of the simulation, vanishes until after 2 seconds, and then reactivates with a low level of activation. During the later phase of balance, these five muscles remain activated, with the rectus femoris showing the highest activation, while the other four muscles not mentioned here remain mostly inactive.

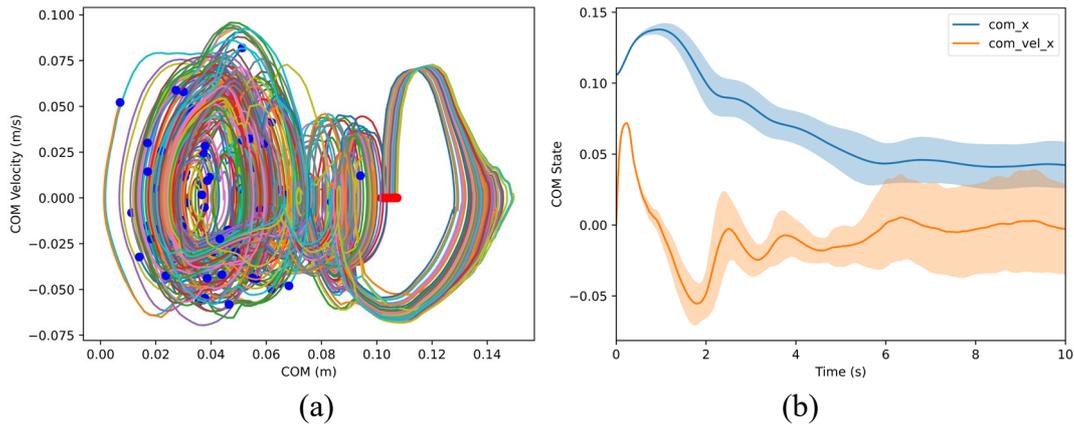

(a)                          (b)

**Figure 7:** (a) COM trajectories of balance recovery from forward lean ( Mean: $8^o$, SD: $0.1^o$ ) using controller trained with method 1 due to its better performance when compared to the other methods. The red points are the starting positions, and the blue points are the end positions at 10 s. For reference, the origin of the COM position is at the same horizontal (x) position as the ankle joint and the x-position of the toe contact point is 0.15 m. (b) The time history profiles of COM state (mean and SD averaged from 100 successful trials). The shaded area displays $\pm SD$ of the mean. $com\_x$ is the COM x position and $com\_vel\_x$ is the COM x velocity.

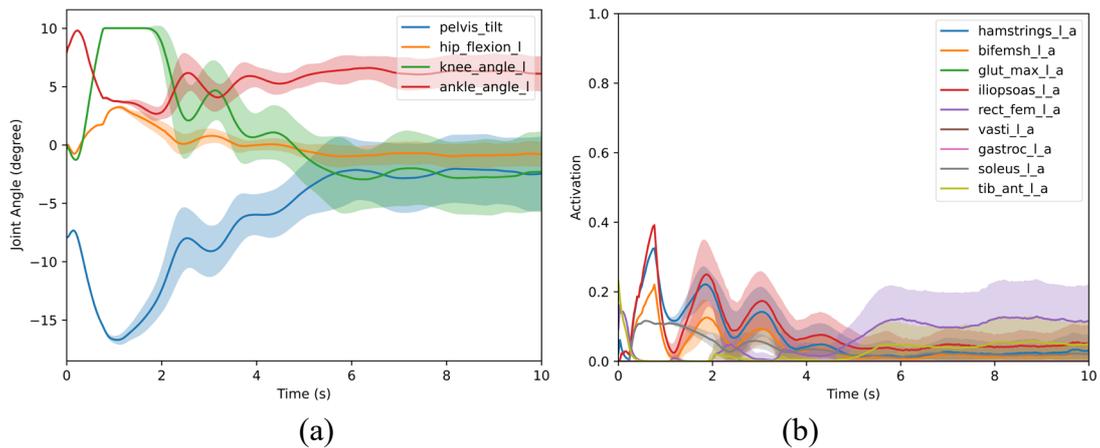

(a)                          (b)

**Figure 8:** Mean and SD plots of (a) joint angles and (b) muscle activations from forward lean recovery. "_l_a" in the legend text indicates muscle activation on the left side.

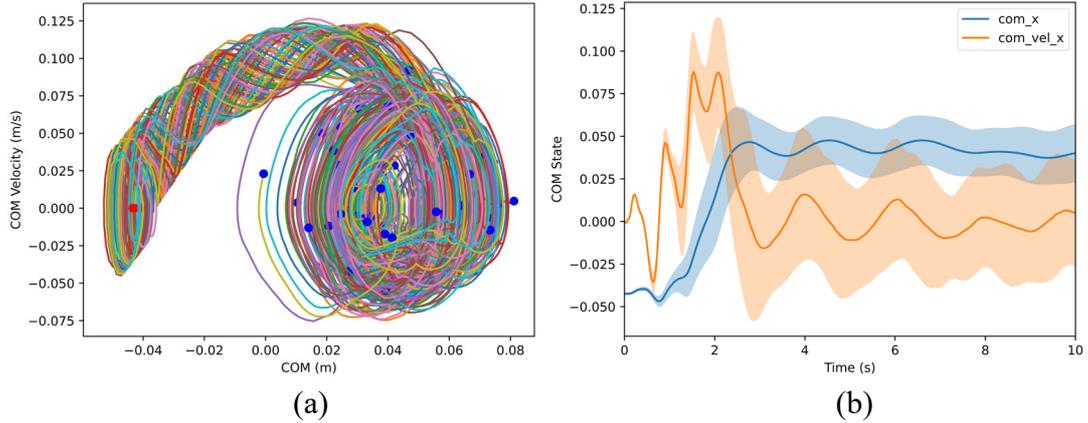

(a)                         (b)

**Figure 9:** (a) COM trajectories of balance recovery from backward lean (Mean: −1.45°, SD: 0.1°). The red points are the starting positions, and the blue points are the end positions at 10 s, where the model is statically stable and within the foot limits. (b) The time history profiles of COM state (mean and SD averaged from 100 successful trials) from backward lean recovery.

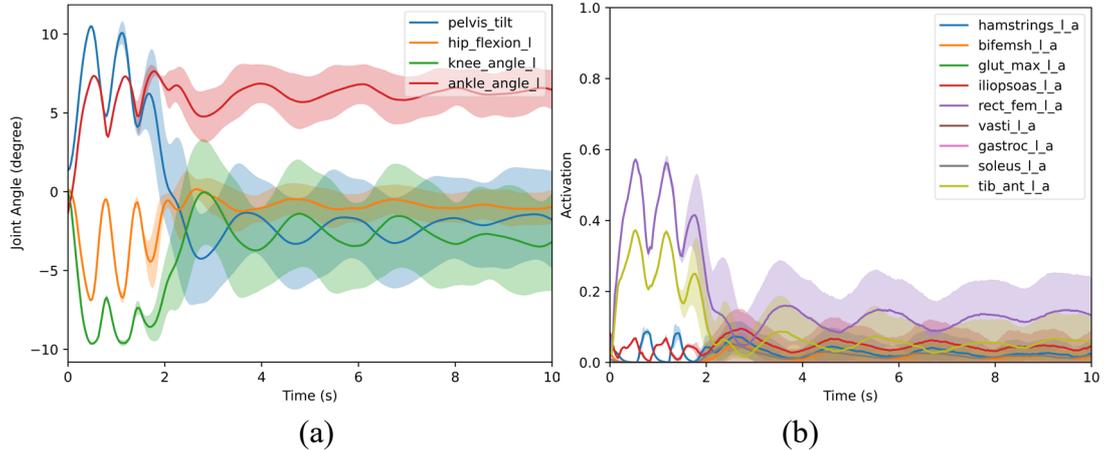

(a)                         (b)

**Figure 10:** Mean and SD plots of (a) joint angles and (b) muscle activations from backward lean recovery.

**Effects of Muscle Weakness and Neural Delay.** Aging and neuromuscular disorders often induce significant changes to muscle physiological properties that affect people's balance. For example, during the process of aging, it has been shown that muscle fiber's maximum isometric force and contraction velocity decreases while activation and deactivation time constants increases [72-74]. To investigate the effect of muscle property degradation on balance recovery, we conducted tests by modifying muscle properties in several ways. In the first case, we reduced the maximum isometric fiber forces of all muscles by 30% to simulate muscle weakness, resembling conditions such as those associated with aging. In the second case, we specifically reduced the maximum isometric fiber forces of muscles on the left side only by 30% (hemiparesis). In the third case, we further reduced the maximum isometric fiber forces of muscles on the left side to 0% of their original strength to simulate complete loss of muscle strength on one side (hemiplegia). In the two latter cases, the maximum isometric fiber forces of muscle on the right side were kept at their original strength. Both cases allowed us to explore the effect of muscle weakness and asymmetry and the last case allowed us to explore the effect of complete muscle disability on one side. Note in the two asymmetric cases, we switched to use a 3D MSK model by changing the 3-DOF planar root joint at the pelvis to a 6-DOF free joint, enabling global lateral movement. In addition, the symmetry condition for the first neural network (CPN) was removed. For each of these cases, we again trained the controller using training method 1 (zero velocity) due to its overall good performance demonstrated earlier and then tested it with random ankle angle and velocity. In Figure 11, we present a comparison of both the PBRs and system BRs for these three cases with muscle weakness on one or both sides. To investigate the effect of muscle activation and deactivation time (Eq. (3)) on the ability to recover balance, we increased both durations by 50% to simulate a longer neuromuscular response time (i.e., neural delay). Subsequently, we conducted the same training process to obtain a new controller and tested it to generate new BRs. Compared to the normal BRs depicted in Figure 5(a), the new BRs are much smaller in terms of the covered area, particularly in the region above the zero-velocity line. This suggests that the controller's performance is compromised when recovering from a large backward inclined angle.

## 4. Discussion

The proposed balance controller trained by the RL framework represents a novel method to explore the limits of dynamic balance of human standing posture, under comprehensive kinematics, contact, and muscle activation constraints. The results demonstrated that effective muscle-based balance controllers can be generated using RL techniques. Our RL approach involved the utilization of two decoupled yet interconnected neural networks, similar to the method employed by Lee et al. [42]; however, unlike their work, we did not rely on any reference motion. Instead, we developed various neuromusculoskeletal physics and balance inspired rewards for controlling balance recovery. These included reaching a target balanced posture, maintaining an upright upper body, and utilizing a LIP model-based balance criterion known as XcoM. We conducted an ablation study by removing selected rewards from the total reward in Eq. (11) and trained additional controllers with method 1. Without the XcoM reward, the generated system BR has an area of 0.0932 with weaker backward lean recovery capability, an overall success rate of 17.29% and a success rate of 51.75% within the BR. Without the upright posture reward, the system BR has an area 0.0706 with weaker forward recovery capability, an overall success rate of 20.46%, and a success rate of 56.77% with the BR. These results signify the importance of these two balance-inspired rewards since including both in the reward drastically increases the success rates (an overall success rate of 59.59% and a success rate of 98.77% within the BR).

To explore the COM state space during balance recovery, we devised a novel procedure (Algorithm 1) for RSI, which was utilized during both training and testing. We also investigated the use of early termination and CL to enhance the efficiency and convergence of balance recovery controllers during training. We found that RSI and early termination were crucial in achieving robust balance recovery controllers, aligning with the observations made by Peng et al [42, 75]. We have experimented to remove some of the early termination conditions, which resulted in the emergence of balance recovery strategies such as foot sliding and stepping when dealing with challenging initial conditions. Notably, when trained without RSI, such as using a fixed initial state, the resulting controller exhibited difficulties in effectively handling unexplored conditions. We employed three distinct training methods, where each utilized different RSI strategies or CL, in order to obtain robust neuromuscular controllers. Among these methods, the controller generated from the first method, which involved starting from a random initial inclination angle with zero velocity, exhibited the best overall performance in terms of robustness (success rate) and coverage of the recoverable COM state space (i.e., BR). In addition to the success rate, the first method's controller also demonstrated more convexity around the static equilibrium (zero-velocity line along the base of support), which is more physically consistent with bipedal systems behaving like inverted pendulums. The CL approach employed in our study did not appear to improve the performance of the controllers. Nonetheless, we believe there are potential avenues for enhancing the CL approach by using more sophisticated training procedures, such as gradually increasing the difficulty of the RSI and adjusting the early termination conditions to relax and then tighten them.

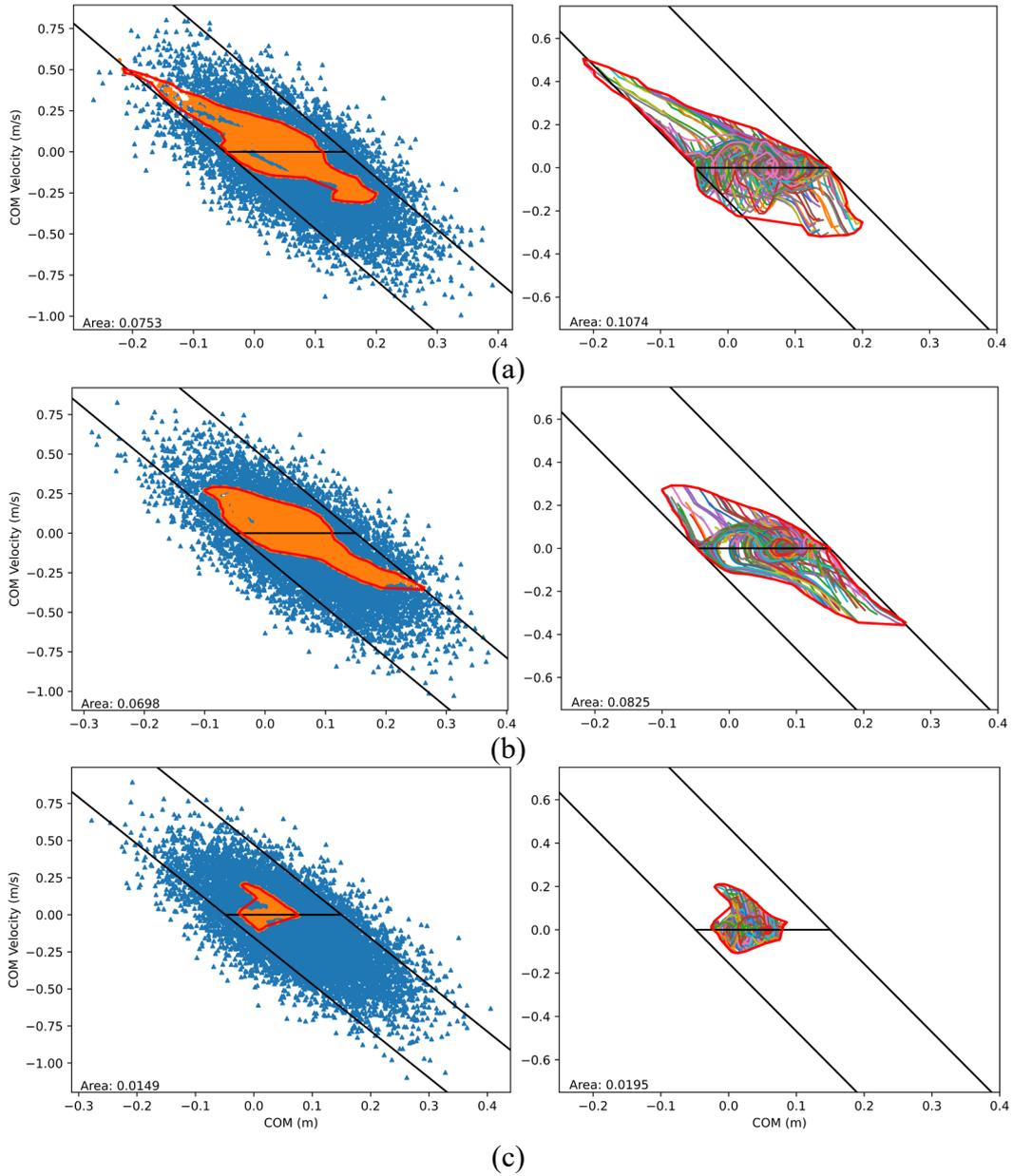

**Figure 11:** COM state space BRs for the learned controllers trained with modified muscle properties. (a) Maximum isometric fiber forces of all muscles were reduced by 30% of their original strength; (b) Maximum isometric fiber forces of muscles on the left side were reduced by 30% of their original strength; (c) Maximum isometric fiber forces of muscles on the left side were reduced to 0% of the original strength for all muscles.

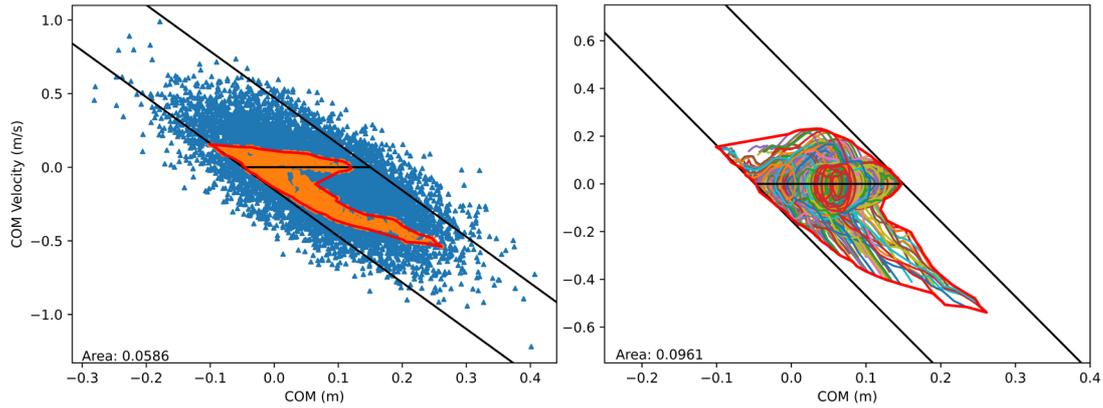

**Figure 12:** COM state space BR for the learned controller trained with muscle activation and deactivation time increased by 50%.

For the time history plots of joint angles and COM positions from Figure 7 to Figure 10, we can observe that the trained models demonstrate a slower response in balance recovery (approximately 3-5 s to reach relative stable states) than that typically observed in humans (approximately 2-3 s based on what we observed from human experiments). This could be due to the higher focus on positional measures than time-related metrics, leading to the model preferring recoveries that enable it to reach the target posture instead of recovering quicker. Additionally, when recovering to its final posture, the RL-trained controllers exhibited an oscillatory effect at maintaining balance rather than steadying itself; this oscillatory effect is also not observed in humans at a large scale. These oscillations could potentially be influenced by the lack of reward that encourages smoothness of the motion as well as the selected stable PD parameters. Further investigation is warranted to understand the specific role played by these factors in contributing to the observed oscillations.

Although there are large overlapping areas in the BRs of the three controllers, none of them encompassed all the areas covered by the other two. This poses a question as to whether the human achievable BR should be the one from the best performing controller or the union of the BRs from all three controllers, the latter of which might be more appropriate if one can train a universal controller that is robust enough. Additionally, if we were to train additional controllers using different training strategies or methods, it would be worthwhile to investigate how much the BR can be expanded, and to what extent or limit, with these controllers.

When comparing the successful COM states shown in the PBRs and the corresponding system BRs generated from the trajectories, we can observe that in general the system BRs cover more COM state space areas than the PBRs. The covered states outside the PBRs are included in the BRs because they reside on the successful balance recovery trajectories. Note, in the RSI, the initial states (at $t = 0$) are only specified for joint angle and angular velocities. However, initial muscle states (i.e., activations) are also very important in determining the balance outcome. One choice is to set muscle activations to zero at the beginning and let it ramp up with the MCN predicted excitation. Since this introduces a delay in muscle response to the initial inclined state, we set the muscle activations to be the same as the excitations to mimic the effect of anticipation. Therefore, two coincident COM states in the PBR and BR could indicate very different overall system dynamic states that consider muscles and result in different balance outcomes. For instance, in Figure 7, it can be observed that the COM state starts within the foot then shifts towards the toe before settling back over the COM of the foot (final balanced state); if the model were to start at the toe, on the other hand, it would likely fall and be counted as an unbalanced COM state because the initial muscle states would not be able to lead to a balanced state at the end.

Ideally, it is desired to have a universal balance controller that is highly robust and capable of recovering from the largest possible area within the COM state space. This objective is likely to require further study employing more advanced training methods, which can effectively explore the control space, avoid convergence to local optima, and utilize sophisticated physical models that account for variations in physics properties and uncertainties in human-environment interaction. In the literature, techniques such as kinematics and dynamics randomization [40, 41, 65, 76-79] have been used to enhance robustness of trained controllers and enable sim-to-real transfer of virtually trained controllers to physical hardware, accounting for modeling inaccuracies or uncertainties in modeling. Similar randomization strategies can potentially enhance the performance of our trained controllers. Besides the randomization of the typical kinematics (e.g., link length, joint positions) and dynamics quantities (e.g., mass, inertia, COM position), we can also randomize muscle properties (e.g., maximum isometric force, activation and deactivation time, maximum contraction velocity, etc.) during the training.

The BRs obtained using various trained RL controllers and the MSK model were largely contained within and aligned with the analytical LIP model-based limits but could not have large COM excursions that deviate too far from the balanced upright posture. This is because of the limits of human strength and joint range of motions, which were not considered in the theoretical model. In the LIP model, since it is for ideal systems, the total area of the analytical limits is infinite in theory, but these limits can then be constrained using the friction cone to obtain a more meaningful BR with appropriate limits [29]. However, the BRs obtained from the RL-based controllers are still comparable because the MSK model was encouraged to focus on recovering its balance using an ankle strategy through the imposed initial states in the RSI, which is similarly aligned with the ankle strategy used in the LIP model.

Although the current method is significantly different from the torque-based motion optimization method, the predicted BRs show comparability with the results obtained from optimization [29]. However, it is worth noting that, in general, the predicted BRs using our muscle controller-based method tend to be smaller. This suggests that the utilization of the MSK model imposes stricter physical and physiological constraints on the feasibility of balance recovery from wider initial states due to muscles' actuation capabilities. The torque-based optimization method does not consider the neural delay and often assumes constant torque capacities during the entire motion, whereas the torques generation capacity of muscles are affected by their states (e.g., fiber lengths) and moment arms at different instantaneous state. Despite of the lack of comprehensive comparison between these two approaches, we believe BRs predicted with MSK models encompass more important physical and physiological factors or constraints affecting balance recovery and are likely to be more realistic from this perspective.

To explore the effects of altered muscle properties on the limits of dynamic balance, we generated RL-based controllers with modified muscle properties, such as maximum isometric fiber forces and activation and deactivation time of the neural excitation-activation delay for all or selected muscles. By analyzing the resulting BRs, we observed that in the first two cases, in which the maximum isometric fiber forces are reduced by 30% on both sides and only one side, the corresponding controllers could still recover from statically balanced states (i.e., COM states with positions within the base of support and zero velocity). Conversely, in the case of fully disabled muscles on one side, the trained controller could not cover the entire set of statically balanced states. This suggests that in pathologies such as hemiplegia, the capability of maintaining a static posture is much reduced, resulting in a much smaller range of feasible COM sway postures that are near the middle of base of support. Additionally, we discovered that asymmetric muscle weakness (30% less strength on one side) produced a smaller BR, particularly in the region of backward inclination, than that of the case with muscle weakness on both sides. It should also be noted that, the asymmetric model has the global translation and rotation DOFs at the pelvis in all directions, but it does not include hip abduction and rotation, considering the muscles included in the MSK model are largely used for actuating motion in the sagittal plane. For the case with longer (50%) activation and deactivation time, the BR is much smaller in the region of backward inclination as well. These findings suggest that individuals or patients with muscle weakness or slower neural response times may have difficulties recovering from backward imbalance and are more prone to falling.

Our numerical experiments demonstrate that these RL-trained muscle controllers have great potential to study human balance in the deeper neuromuscular domain and provide valuable insights on the factors that influence balance improvement or deterioration. The obtained BRs under different muscle conditions offer valuable information regarding the capabilities and limitations of balance recovery of individuals or patients with symptoms such as muscle weakness or hemiplegia. These findings can also be relevant for fall detection and monitoring of the margin of stability during daily activities, especially if a personalized BR can be established through subject-specific modeling and control. Our work can be extended to study balance in other patient populations, such as cerebral palsy patients and people with Parkinson's disease. By adapting the RL-trained muscle controllers to these specific conditions, we can gain further insights into balance challenges faced by these individuals.

## 5. Conclusions

An RL framework was developed to effectively learn muscle-based balance controllers and utilized them to establish physiologically feasible regions of stability for standing balance (i.e., the BRs). Several key factors have contributed to the novel outcomes of this work. Firstly, neuromusculoskeletal physics and balance-inspired rewards, as well as balance-related initial states, were incorporated in the RL training. This ensured that the controllers learned to prioritize actions that align with the principles of human balance during standing. Secondly, two separate yet interconnected neural networks were utilized that separately generated control policies for torques and muscle activations, which was expected to improve performance based on the results from [42]. Additionally, a combination of training strategies including novel RSI, early termination, and curriculum learning were employed to test the efficiency and effectiveness of the learning processes. To evaluate the performance of the trained controllers, we compared them based on different training strategies and examined the resulting BRs under various muscle conditions. By comparing the obtained BRs with the theoretical limits of dynamic balance defined by the LIP model, we gained valuable insights into human balance recovery by considering the physiological capabilities and limitations of the human musculoskeletal system. This study has laid a solid foundation for the stability region-based analysis of human balance, integrating physiological factors and whole-body biomechanics. This approach surpasses traditional methods of balance control and assessment that rely on reduced-order kinematics or ground reference points and provides more informative and subject-specific limits of dynamic balance.

Moving forward, the robustness of the learned controllers will be enhanced by introducing domain randomization, perturbations, and more sophisticated learning or training methods. This will improve their adaptability to a wider range of balance scenarios. Additionally, we aim to explore more balance-specific reward formulations to further optimize the training process and improve the assessment of balance. Furthermore, an in-depth analysis of the resulting balance kinematics and muscle activations predicted by this RL framework will be conducted, through a comparison with empirical data collected from postural sway exercises in a laboratory setting. Future experimental investigation will include the collection of both kinematic measurements and muscle EMG recordings from selected relevant lower-limb muscles. By combining simulation-based analysis with experimental validation, we can refine and validate the RL framework's predictions, making it a more reliable tool for studying human balance and its underlying mechanisms.

**References**

1. Gandolfi S, Carloni R, Bertheuil N, Grolleau JL, Auquit-Auckbur I, Chaput B. Assessment of quality-of-life in patients with face-and-neck burns: The Burn-Specific Health Scale for Face and Neck (BSHS-FN). Burns. 2018;44(6):1602-9. Epub 20180627. doi: 10.1016/j.burns.2018.03.002. PubMed PMID: 29958746.

2. Levinger P, Dunn J, Bifera N, Butson M, Elias G, Hill KD. High-speed resistance training and balance training for people with knee osteoarthritis to reduce falls risk: study protocol for a pilot randomized controlled trial. Trials. 2017;18(1):384. Epub 20170818. doi: 10.1186/s13063-017-2129-7. PubMed PMID: 28821271; PubMed Central PMCID: PMCPMC5563024.

3. Visser JE, Carpenter MG, van der Kooij H, Bloem BR. The clinical utility of posturography. Clinical Neurophysiology. 2008;119(11):2424-36. Epub 20080912. doi: 10.1016/j.clinph.2008.07.220. PubMed PMID: 18789756.
4. Jager TE, Weiss HB, Coben JH, Pepe PE. Traumatic Brain Injuries Evaluated in U.S. Emergency Departments, 1992-1994. Academic Emergency Medicine. 2000;7(2):134-40. doi: 10.1111/j.1553-2712.2000.tb00515.x.
5. Stevenson TJ. Detecting change in patients with stroke using the Berg Balance Scale. Australian Journal of Physiotherapy. 2001;47(1):29-38. doi: https://doi.org/10.1016/S0004-9514(14)60296-8.
6. Godi M, Franchignoni F, Caligari M, Giordano A, Turcato AM, Nardone A. Comparison of reliability, validity, and responsiveness of the mini-BESTest and Berg Balance Scale in patients with balance disorders. Physical Therapy. 2013;93(2):158-67. Epub 20120927. doi: 10.2522/ptj.20120171. PubMed PMID: 23023812.
7. Bell DR, Guskiewicz KM, Clark MA, Padua DA. Systematic review of the balance error scoring system. Sports Health. 2011;3(3):287-95. doi: 10.1177/1941738111403122. PubMed PMID: 23016020; PubMed Central PMCID: PMCPMC3445164.
8. Raad J, Moore J, Hamby J, Rivadelo RL, Straube D. A Brief Review of the Activities-Specific Balance Confidence Scale in Older Adults. Archives of Physical Medicine and Rehabilitation. 2013;94(7):1426-7. doi: 10.1016/j.apmr.2013.05.002.
9. Kinzey SJ, Armstrong CW. The reliability of the star-excursion test in assessing dynamic balance. Journal of Orthopaedic and Sports Physical Therapy. 1998;27(5):356-60. doi: 10.2519/jospt.1998.27.5.356. PubMed PMID: 9580895.
10. Glave AP, Didier JJ, Weatherwax J, Browning SJ, Fiaud V. Testing Postural Stability: Are the Star Excursion Balance Test and Biodex Balance System Limits of Stability Tests Consistent? Gait and Posture. 2016;43:225-7. Epub 20151026. doi: 10.1016/j.gaitpost.2015.09.028. PubMed PMID: 26514832.
11. Tinetti ME. Performance-Oriented Assessment of Mobility Problems in Elderly Patients. Journal of the American Geriatrics Society. 1986;34(2):119-26. doi: https://doi.org/10.1111/j.1532-5415.1986.tb05480.x.
12. Ruhe A, Fejer R, Walker B. Center of pressure excursion as a measure of balance performance in patients with non-specific low back pain compared to healthy controls: a systematic review of the literature. European Spine Journal. 2011;20(3):358-68. doi: 10.1007/s00586-010-1543-2.
13. Lin D, Seol H, Nussbaum MA, Madigan ML. Reliability of COP-based postural sway measures and age-related differences. Gait & Posture. 2008;28(2):337-42. doi: https://doi.org/10.1016/j.gaitpost.2008.01.005.
14. O'Connor SM, Baweja HS, Goble DJ. Validating the BTrackS Balance Plate as a low cost alternative for the measurement of sway-induced center of pressure. Journal of Biomechanics. 2016;49(16):4142-5. doi: https://doi.org/10.1016/j.jbiomech.2016.10.020.
15. Menant JC, Latt MD, Menz HB, Fung VS, Lord SR. Postural sway approaches center of mass stability limits in Parkinson's disease. Movement Disorders. 2011;26(4):637-43. doi: https://doi.org/10.1002/mds.23547.
16. Yeung LF, Cheng KC, Fong CH, Lee WCC, Tong K-Y. Evaluation of the Microsoft Kinect as a clinical assessment tool of body sway. Gait & Posture. 2014;40(4):532-8. doi: https://doi.org/10.1016/j.gaitpost.2014.06.012.
17. Doheny EP, McGrath D, Greene BR, Walsh L, McKeown D, Cunningham C, et al., editors. Displacement of centre of mass during quiet standing assessed using accelerometry in older fallers and non-fallers. 2012 Annual International Conference of the IEEE Engineering in Medicine and Biology Society; 2012 28 Aug.-1 Sept. 2012.
18. Peng WZ, Song H, Kim JH. Stability Region-Based Analysis of Walking and Push Recovery Control. Journal of Mechanisms and Robotics. 2021;13(3):1-11. doi: 10.1115/1.4050095.
19. Peng WZ, Mummolo C, Song H, Kim JH. Whole-body balance stability regions for multi-level momentum and stepping strategies. Mechanism and Machine Theory. 2022;174. doi: 10.1016/j.mechmachtheory.2022.104880.
20. Mummolo C, Mangialardi L, Kim JH. Numerical Estimation of Balanced and Falling States for Constrained Legged Systems. Journal of Nonlinear Science. 2017;27:1291-323. doi: 10.1007/s00332-016-9353-2.
21. Stephens BJ, Atkeson CG, editors. Push Recovery by stepping for humanoid robots with force controlled joints2010 2010: IEEE.
22. Yang C, Wu Q. Effects of constraints on bipedal balance control during standing. International Journal of Humanoid Robotics. 2007;4(4):753-75. doi: 10.1109/acc.2006.1656599.
23. Wieber PB, editor On the stability of walking systems. Proceedings of the International Workshop on Humanoid and Human Friendly Robotics; 2002; Tsukuba, Japan.
24. Zaytsev P, Hasaneini SJ, Ruina A. Two steps is enough: No need to plan far ahead for walking balance. Proceedings - IEEE International Conference on Robotics and Automation: IEEE; 2015. p. 6295-300.
25. Koolen T, de Boer T, Rebula J, Goswami A, Pratt J. Capturability-based analysis and control of legged locomotion, Part 1: Theory and application to three simple gait models. The International Journal of Robotics Research. 2012;31(9):1094-113. doi: 10.1177/0278364912452673.
26. Mummolo C, Peng WZ, Agarwal S, Griffin R, Neuhaus PD, Kim JH. Stability of Mina V2 for robot-assisted balance and locomotion. Frontiers in Neurorobotics2018. p. 1-16.
27. Mummolo C, Peng WZ, Gonzalez C, Kim JH. Contact-dependent balance stability of biped robots. Journal of Mechanisms and Robotics2018. p. 1-13.
28. Akbas K, Mummolo C. A Computational Framework Towards the Tele-Rehabilitation of Balance Control Skills. Frontiers in Robotics and AI. 2021;8:648485. Epub 20210609. doi: 10.3389/frobt.2021.648485. PubMed PMID: 34179106; PubMed Central PMCID: PMCPMC8220374.
29. Mummolo C, Akbas K, Carbone G. State-Space Characterization of Balance Capabilities in Biped Systems with Segmented Feet. Frontiers in Robotics and AI. 2021;8:613038. Epub 20210226. doi: 10.3389/frobt.2021.613038. PubMed PMID: 33718440; PubMed Central PMCID: PMCPMC7952635.
30. Winters J. Hill-Based Muscle Models: A Systems Engineering Perspective. 1990. p. 69-93.
31. Winters JM, Stark L. Muscle models: What is gained and what is lost by varying model complexity. Biological Cybernetics. 1987;55(6):403-20. doi: 10.1007/BF00318375.
32. Layne CS, Malaya CA, Ravindran AS, John I, Francisco GE, Contreras-Vidal JL. Distinct Kinematic and Neuromuscular Activation Strategies During Quiet Stance and in Response to Postural Perturbations in Healthy Individuals Fitted With and Without a Lower-Limb Exoskeleton. Frontiers in Human Neuroscience. 2022;16. doi: 10.3389/fnhum.2022.942551.

33. McKay JL, Lang KC, Bong SM, Hackney ME, Factor SA, Ting LH. Abnormal center of mass feedback responses during balance: A potential biomarker of falls in Parkinson's disease. PLoS One. 2021;16(5):e0252119. Epub 20210527. doi: 10.1371/journal.pone.0252119. PubMed PMID: 34043678; PubMed Central PMCID: PMCPMC8158870.
34. Romanato M, Volpe D, Guiotto A, Spolaor F, Sartori M, Sawacha Z. Electromyography-informed modeling for estimating muscle activation and force alterations in Parkinson's disease. Computer Methods in Biomechanics and Biomedical Engineering. 2022;25(1):14-26. doi: 10.1080/10255842.2021.1925887.
35. Thelen DG, Anderson FC, Delp SL. Generating dynamic simulations of movement using computed muscle control. Journal of biomechanics. 2003;36(3):321--8. PubMed PMID: Thelen2003.
36. Zhou X, Chen X. Design and Evaluation of Torque Compensation Controllers for a Lower Extremity Exoskeleton. Journal of Biomechanical Engineering. 2021;143(1):11.
37. Weng J, Hashemi E, Arami A. Natural walking with musculoskeletal models using deep reinforcement learning. IEEE Robotics and Automation Letters. 2021;6(2):4156-62.
38. Song S, Kidziński Ł, Peng XB, Ong C, Hicks J, Levine S, et al. Deep reinforcement learning for modeling human locomotion control in neuromechanical simulation. Journal of neuroengineering and rehabilitation. 2021;18:1-17.
39. Kidziński Ł, Ong C, Mohanty SP, Hicks J, Carroll S, Zhou B, et al., editors. Artificial intelligence for prosthetics: Challenge solutions. The NeurIPS'18 Competition: From Machine Learning to Intelligent Conversations; 2020: Springer.
40. Luo S, Androwis G, Adamovich S, Nunez E, Su H, Zhou X. Robust Walking Control of a Lower Limb Rehabilitation Exoskeleton Coupled with a Musculoskeletal Model via Deep Reinforcement Learning. Journal of NeuroEngineering and Rehabilitation. 2023;20(34). doi: https://doi.org/10.1186/s12984-023-01147-2.
41. Luo S, Androwis G, Adamovich S, Su H, Nunez E, Zhou X. Reinforcement Learning and Control of a Lower Extremity Exoskeleton for Squat Assistance. Frontiers in Robotics and AI. 2021;8.
42. Lee S, Park M, Lee K, Lee J. Scalable muscle-actuated human simulation and control. ACM Transactions on Graphics. 2019;38. doi: 10.1145/3306346.3322972.
43. Kober J, Bagnell JA, Peters J. Reinforcement learning in robotics: A survey. International Journal of Robotics Research. 2013;32:1238-74. doi: 10.1177/0278364913495721.
44. Wang S, Chaovalitwongse W, Babuška R. Machine learning algorithms in bipedal robot control. IEEE Transactions on Systems, Man and Cybernetics Part C: Applications and Reviews: IEEE; 2012. p. 728-43.
45. Kumar VCV, Ha S, Sawicki G, Liu CK, editors. Learning a Control Policy for Fall Prevention on an Assistive Walking Device2020 2020: IEEE.
46. Bogdanovic M, Khadiv M, Righetti L. Model-free Reinforcement Learning for Robust Locomotion using Demonstrations from Trajectory Optimization. Frontiers in Robotics and AI. 2022;9. doi: 10.3389/frobt.2022.854212.
47. Xie Z, Berseth G, Clary P, Hurst J, Van De Panne M, editors. Feedback Control For Cassie With Deep Reinforcement Learning. International Conference on Intelligent Robots and Systems (IROS); 2018; Madrid, Spain: IEEE.
48. Yang C, Komura T, Li Z. Emergence of human-comparable balancing behaviours by deep reinforcement learning. IEEE-RAS International Conference on Humanoid Robots. 2017:372-7. doi: 10.1109/HUMANOIDS.2017.8246900.

49. Li T, Geyer H, Atkeson CG, Rai A. Using deep reinforcement learning to learn high-level policies on the ATRIAS biped. Proceedings - IEEE International Conference on Robotics and Automation. 2019;2019-May:263-9. doi: 10.1109/ICRA.2019.8793864.
50. Lin JL, Hwang KS, Jiang WC, Chen YJ. Gait Balance and Acceleration of a Biped Robot Based on Q-Learning. IEEE Access. 2016;4:2439-49. doi: 10.1109/ACCESS.2016.2570255.
51. Wu W, Gao L. Posture self-stabilizer of a biped robot based on training platform and reinforcement learning. Robotics and Autonomous Systems. 2017;98:42-55. doi: 10.1016/j.robot.2017.09.001.
52. Joe HM, Oh JH. Balance recovery through model predictive control based on capture point dynamics for biped walking robot. Robotics and Autonomous Systems. 2018;105:1-10. doi: 10.1016/j.robot.2018.03.004.
53. Wang S, Wang L, Meijneke C, Van Asseldonk E, Hoellinger T, Cheron G, et al. Design and Control of the MINDWALKER Exoskeleton. IEEE Transactions on Neural Systems and Rehabilitation Engineering. 2015;23:277-86. doi: 10.1109/TNSRE.2014.2365697.
54. Xi A, Chen C. Stability Control of a Biped Robot on a Dynamic Platform Based on Hybrid Reinforcement Learning. Sensors. 2020;20(16):4468. doi: 10.3390/s20164468.
55. Delp SL, Anderson FC, Arnold AS, Loan P, Habib A, John CT, et al. OpenSim: open-source software to create and analyze dynamic simulations of movement. IEEE Transactions on Biomedical Engineering. 2007;54(11):1940-50. doi: 10.1109/TBME.2007.901024. PubMed PMID: 18018689.
56. Millard M, Uchida T, Seth A, Delp SL. Flexing computational muscle: modeling and simulation of musculotendon dynamics. Journal of biomechanical engineering. 2013;135(2).
57. Zajac FE. Muscle and tendon: properties, models, scaling, and application to biomechanics and motor control. Critical reviews in biomedical engineering. 1989;17(4):359--411. PubMed PMID: Zajac1989.
58. Mousavi A, Ehsani H, Rostami M, editors. Compliant Vs. rigid tendon models: a simulation study on precision, computational efficiency and numerical stability. 2014 21th Iranian Conference on Biomedical Engineering (ICBME); 2014: IEEE.
59. Todorov E, Erez T, Tassa Y, editors. Mujoco: A physics engine for model-based control. 2012 IEEE/RSJ international conference on intelligent robots and systems; 2012: IEEE.
60. Lee J, Grey MX, Ha S, Kunz T, Jain S, Ye Y, et al. DART: Dynamic Animation and Robotics Toolkit. Journal of Open Source Software. 2018;3(22).
61. Tan J, Liu K, Turk G. Stable proportional-derivative controllers. IEEE Computer Graphics and Applications. 2011;31(4):34-44.
62. Schulman J, Wolski F, Dhariwal P, Radford A, Klimov O. Proximal Policy Optimization Algorithms. ArXiv. 2017.
63. Hof AL, Gazendam MGJ, Sinke WE. The condition for dynamic stability. Journal of Biomechanics. 2005;38:1-8. doi: 10.1016/j.jbiomech.2004.03.025.
64. Peng XB, Abbeel P, Levine S, Van de Panne M. Deepmimic: Example-guided deep reinforcement learning of physics-based character skills. ACM Transactions On Graphics (TOG). 2018;37(4):1-14.

65. Tan J, Zhang T, Coumans E, Iscen A, Bai Y, Hafner D, et al., editors. Sim-to-Real: Learning Agile Locomotion For Quadruped Robots. Robotics: Science and Systems; 2018; Pittsburgh, Pennsylvania.
66. Bengio Y, Louradour J, Collobert R, Weston J, editors. Curriculum learning. Proceedings of the 26th annual international conference on machine learning; 2009.
67. Xie Z, Ling HY, Kim NH, van de Panne M, editors. Allsteps: curriculum‐driven learning of stepping stone skills. Computer Graphics Forum; 2020: Wiley Online Library.
68. Lee J, Hwangbo J, Wellhausen L, Koltun V, Hutter M. Learning quadrupedal locomotion over challenging terrain. Science robotics. 2020;5(47):eabc5986.
69. Edelsbrunner H, Kirkpatrick DG, Seidel R. On the shape of a set of points in the plane. IEEE Transactions on Information Theory. 1983;29(4):551-9. doi: 10.1109/TIT.1983.1056714.
70. Seth A, Hicks JL, Uchida TK, Habib A, Dembia CL, Dunne JJ, et al. OpenSim: Simulating musculoskeletal dynamics and neuromuscular control to study human and animal movement. PLOS Computational Biology. 2018;14(7):e1006223. doi: 10.1371/journal.pcbi.1006223.
71. Hamner SR, Seth A, Delp SL. Muscle contributions to propulsion and support during running. Journal of Biomechanics. 2010;43(14):2709-16. doi: https://doi.org/10.1016/j.jbiomech.2010.06.025.
72. Cseke B. Simulating Ideal Assistive Devices to Reduce the Metabolic Cost of Walking in the Elderly [Thesis]. Ottawa, Ontario, Canada: University of Ottawa; 2020.
73. Doherty TJ, Vandervoort AA, Brown WF. Effects of Ageing on the Motor Unit: A Brief Review. Canadian Journal of Applied Physiology. 1993;18(4):331-58. doi: 10.1139/h93-029 %M 8275048.
74. Thelen DG. Adjustment of Muscle Mechanics Model Parameters to Simulate Dynamic Contractions in Older Adults. Journal of Biomechanical Engineering. 2003;125(1):70-7. doi: 10.1115/1.1531112.
75. Peng XB, Abbeel P, Levine S, Panne Mvd. DeepMimic: example-guided deep reinforcement learning of physics-based character skills. ACM Transactions on Graphics. 2018;37(4):Article 143. doi: 10.1145/3197517.3201311.
76. Exarchos I, Jiang Y, Yu W, Karen Liu C. Policy Transfer via Kinematic Domain Randomization and Adaptation.  2021 IEEE International Conference on Robotics and Automation (ICRA)2021. p. 45-51.
77. Vinitsky E, Du Y, Parvate KV, Jang K, Abbeel P, Bayen A. Robust Reinforcement Learning Using Adversarial Populations.  International Conference on Learning Representations; Virtual2021.
78. Rajeswaran A, Ghotra S, Ravindran B, Levine S. EPOPT: Learning Robust Neural Network Policies Using Model Ensembles.  International Conference on Learning Representations; Toulon, France2017.
79. Ding Z, Dong H. Challenges of Reinforcement Learning. In: Dong H, Ding Z, Zhang S, editors. Deep Reinforcement Learning: Fundamentals, Research and Applications. Singapore: Springer Singapore; 2020. p. 249-72.